%% file: bfkcorr.tex
\documentclass[11pt]{article}
\newcommand{\play}{\mathit{play}}
\newcommand{\RAT}{\mathit{RAT}}
\newcommand{\strat}{\mathbf{s}}
\newcommand{\intension}[1]{[\![ #1 ]\!]_M}
\newcommand{\intensionc}[1]{[\![ #1 ]\!]_{M^c}}
\newcommand{\intensioncp}[1]{[\![ #1 ]\!]_{M^c}'}
\newcommand{\dbi}{\langle B_i \rangle}
\newcommand{\dbj}{\langle B_j \rangle}
\newcommand{\PR}{\mathcal{PR}}
\renewcommand{\L}{\mathcal{L}}
\newcommand{\pr}{\mathit{pr}}
\renewcommand{\S}{\Sigma}

\usepackage{amssymb}
\usepackage{chicagor}
\usepackage{times}
\input defn


\renewcommand{\Z}{{\cal Z}}
\newcommand{\undominated}{\mathit{undominated}}
\newcommand{\shortv}{\commentout}
\newcommand{\fullv}[1]{#1}
\shortv{\linespread{1.02}}

\begin{document}

\begin{titlepage}
\title{A Logical Characterization of Iterated Admissibility}

\author{Joseph Y. Halpern and Rafael Pass\\
Computer Science Department, Cornell
University, Ithaca, NY, 14853, U.S.A. \\  e-mail:
halpern@cs.cornell.edu, rafael@cs.cornell.edu
}

\maketitle

\begin{abstract}
Brandenburger, Friedenberg, and Keisler provide an epistemic
characterization of iterated admissibility (i.e., iterated deletion of
weakly dominated strategies) where uncertainty is represented using LPSs
(lexicographic probability sequences).  Their characterization holds in
a rich structure called a \emph{complete} structure, where all types are
possible.  Here, a logical charaacterization of iterated admisibility
is given that involves only standard probability and holds in all
structures, not just complete structures.  A stronger notion of \emph{strong
admissibility} is then defined.  Roughly speaking, strong
admissibility is meant to capture the intuition that ``all the agent knows''
is that the other agents satisfy the appropriate rationality
assumptions.  Strong admissibility makes it possible to relate
admissibility, \emph{canonical} structures (as typically considered in
completeness proofs in modal logic), complete structures, and the notion
of ``all I know''. 
\end{abstract}
\thispagestyle{empty}
\end{titlepage}

\section{Introduction}
\emph{Admissibility}  is an old criterion in decision making.  A
strategy for player $i$ is admissible if it is a best response to some
belief of player $i$ that puts positive probability on all the
strategy profiles for the other players.  Part of the interest in
admissibility comes from the observation (due to Pearce
\citeyear{Pearce84}) that a strategy $\sigma$ for
player $i$ is admissible iff it is not weakly dominated; that is, there
is no strategy $\sigma'$ for player $i$ that gives $i$ at least as high
a payoff as $\sigma$ no matter what strategy the other players are
using, and sometimes gives $i$ a higher payoff. 

It seems natural to ignore strategies that are not admissible.  But there
is a conceptual problem when it comes to dealing with \emph{iterated}
admissibility (i.e., iterated deletion of weaklhy dominated
strategies).  As Mas-Colell, Whinston, and Green
\citeyear[p. 240]{MWG95} put  
\fullv{in their textbook when discussing iterated deletion of weakly dominated
strategies:} 
\shortv{it:}
\begin{quote}
[T]he argument for deletion of a weakly dominated strategy for player
$i$ is that he contemplates the possibility that every strategy
combination of his rivals occurs with positive probability.  However,
this hypothesis clashes with the logic of iterated deletion, which
assumes, precisely, that eliminated strategies are not expected to
occur.
\end{quote}

Brandenburger, Friedenberg, and Keisler \citeyear{BFK04} (BFK from now
on) resolve this
paradox in the context of iterated deletion of weakly dominated strategies
by assuming that strategies are not really eliminated.  Rather, they
assumed that strategies that are weakly dominated occur with
infinitesimal (but nonzero) probability.   (Formally, this is captured
by using an LPS---\emph{lexicographically ordered probability
sequence}.)  They define a notion of belief (which they call
\emph{assumption}) appropriate for their setting, and show that
strategies that survive $k$ rounds of iterated deletion are ones that
are played in states where there there is $k$th-order mutual belief in
rationality; that is, everyone assume that everyone assumes \ldots ($k-1$
times) that everyone is rational.  However, they prove only that their
characterization of iterated admissibility holds in particularly rich
structures called \emph{complete} structures (defined formally in
Section~\ref{sec:complete}), where all types are 
possible.

Here, we provide an alternate logical characterization of iterated
admissibility. 
The characterization simply formalizes the intuition that an
agent must consider possible all strategies consistent with the rationality
assumptions he is making.   Repeated iterations correspond to stronger
rationality asumptions. The characterization has the advantage that it
holds in all structures, not just complete structures, and assumes  
that agents represent their uncertainty using standard probability
meaures, rather than LPS's or nonstandard probability measures (as is
done in a characterization of Rajan \citeyear{Rajan98}).  Moreover, 
while complete structures must be uncountable, we show that our
characterization is always satisfible in a structure with finitely many
states.

In an effort to understand better the role of complete structures, we
consider \emph{strong admissibility}.  Roughly speaking, strong
admissibility is meant to capture the intuition that ``all the agent knows''
is that the other agents satisfy the appropriate rationality
assumptions.  We are using the phrase ``all agent $i$ knows'' here in
the same 
sense that it is used by Levesque \citeyear{Lev5} and Halpern and
Lakemeyer \citeyear{HalLak94}.  We formalize strong admissibility by 
requiring that the agent ascribe positive probability to all formulas
consistent with his rationality assumptions.  (This admittedly fuzzy
description is made precise in Section~\ref{sec:strong}.) We give a
logical characterization of iterated strong admissibility and show that
a strategy $\sigma$ survives iterated deletion of weakly dominated strategies
iff there is a structure and a state where $\sigma$ is played and
the formula characterizing iterated strong admissibility holds.  While
we can take the structure where the formula holds to be countable,
perhaps the most natural structure to consider is the \emph{canonical}
structure, which has a state corresponding to very satisfiable collection of
formulas.  The canonical structure is uncountable.  

We can show that the canonical structure is complete in the sense of
BFK.  Moreover, under a technical assumption, every complete structure
is essentially canonical (i.e., it has a state corresponding to every
satisfiable collection of formulas).  This sequence of results allows us
to connect (iterated admissibility), complete structures, canonical
structures, and the notion of ``all I know''.

\section{Characterizing Iterated Deletion}

We consider normal-form games with $n$ players.  
Given a (normal-form) $n$-player game $\Gamma$, let
$\Sigma_i(\Gamma)$ 
denote the strategies of player $i$ in $\Gamma$.  We omit the
parenthetical $\Gamma$ when it is clear from context or irrelevant.  Let
$\vec{\Sigma} = \Sigma_1 \times \cdots \times \Sigma_n$.

Let $\L_1$ be the language where we start with $\true$ and the special
primitive 
proposition $\RAT_i$ and close off under modal operators $B_i$ and
$\dbi$, for $i = 1, \ldots, n$, conjunction, and negation.   
We think of $B_i \phi$ as saying that $\phi$ holds with probability 1, and
$\dbi \phi$ as saying that $\phi$ holds with positive probability.
As we shall see, $\dbi$ is definable as $\neg B_i \neg$ if we make the
appropriate measurability assumptions.

To reason about the game $\Gamma$,
we consider a class of probability structures corresponding to
$\Gamma$.
A \emph{probability structure $M$ appropriate for $\Gamma$} 
is a tuple 
$(\Omega,\strat,\F, \PR_1, \ldots, \PR_n)$, where
$\Omega$ is a set of  
states; $\strat$ associates with each state $\omega \in \Omega$ a
pure strategy profile $\strat(\omega)$ in the game $\Gamma$; 
$\F$ is a $\sigma$-algebra over $\Omega$; and, for each
player $i$, $\PR_i$ associates with each state $\omega$ a probability
distribution 
$\PR_i(\omega)$ on $(\Omega,\F)$ 
such that, (1) for each strategy $\sigma_i$ for player $i$,
$\intension{\sigma_i} = \{\omega: \strat_i(\omega) = \sigma_i\} \in \F$, 
where $\strat_i(\omega)$ denotes player $i$'s strategy in the strategy
profile $\strat(\omega)$; (2)
$\PR_i(\omega)(\intension{\strat_i(\omega)}) 
= 1$;
(3) for each probability
measure $\pi$ on $(\Omega, \F)$, and player $i$, $\intension{\pi,i}
=\{\omega : \Pi_i(\omega) = \pi\} \in \F$; and (4) 
$\PR_i(\omega)(\intension{\PR_i(\omega),i}) = 1$.
These assumptions essentially say that player $i$ knows his strategy and
knows his beliefs.

The semantics is given as follows:  
\begin{itemize}
\item $(M,\omega) \sat \true$ (so $\true$ is vacuously true).
\item $(M,\omega) \sat \RAT_i$ if $\strat_i(\omega)$ is a best response,
given player $i$'s beliefs on the strategies of other players induced by
$\PR_i(\omega)$.  (Because we restrict to appropriate structures, a
players expected utility at a state $\omega$ is well defined, so we can
talk about best responses.)
\item $(M,\omega) \sat \neg \phi$ if $(M,\omega) \not\sat
\phi$.  
\item $(M,\omega) \sat \phi \land \phi'$ iff $(M,\omega) \sat \phi$ and 
$(M,\omega) \sat \phi'$
\item $(M,\omega) \sat B_i \phi$ if there exists a set $F \in \F_i$ such
that $F \subseteq \intension{\phi}$ and $\PR_i(\omega)(F) = 1$, where
$\intension{\phi} = \{\omega: (M,\omega) \sat \phi\}$.
\item $(M,\omega) \sat \dbi \phi$ if there exists a set $F \in \F_i$ such
that $F \subseteq \intension{\phi}$ and $\PR_i(\omega)(F) > 0$.
\end{itemize}

Given a language (set of formulas) $\L$, $M$   is
\emph{$\L$-measurable} if $M$ is appropriate (for some game $\Gamma$) and 
$\intension{\phi} \in \F$ for all formulas $\phi \in \L$.  It is easy to
check that in an $\L_1$-measurable structure, $\dbi \phi$ is equivalent
to $\neg B_i \neg \phi$.

To put our results on iterated admissibility into context, we first
consider rationalizability.
Pearce \citeyear{Pearce84} gives two definitions of
rationalizability, which give rise to different epistemic
characterizations.  We repeat the definitions here,
using the notation of Osborne and Rubinstein \citeyear{OR94}.

\dfn\label{rat1} A strategy
$\sigma$ for player~$i$ in game~$\Gamma$  
is \emph{rationalizable} if, for each player $j$, there is a set $\Z_j   
\subseteq \Sigma_j(\Gamma)$ and, for each strategy $\sigma' \in \Z_j$,  a   
probability measure $\mu_{\sigma'}$ on $\Sigma_{-j}(\Gamma)$ whose
support is a subset of
$\Z_{-j}$ such that     
\fullv{
\begin{itemize}   
\item $\sigma \in \Z_i$; and}
\shortv{$\sigma \in \Z_i$ and,}
\fullv{\item} for each player $j$ and strategy $\sigma' \in \Z_j$, 
strategy $\sigma'$ is a best response to (the beliefs) $\mu_{\sigma'}$.   
\fullv{\end{itemize}}
\edfn

The second definition characterizes rationalizability in terms of
iterated deletion.
\dfn\label{rat2} A strategy
$\sigma$ for player~$i$ in game~$\Gamma$  
is \emph{rationalizable$'$} if, for each player $j$, there exists a
sequence $X_j^0, X_j^1, X_j^2, \ldots$ of sets of strategies for player
$j$ such that $X_j^0 = \Sigma_j$ and, for each strategy $\sigma' \in
X_j^k$, $k \ge 1$, a probability measure $\mu_{\sigma',k}$ whose support
is a subset of 
$\vec{X}_{-j}^{k-1}$ such that 
\fullv{\begin{itemize}   
\item $\sigma \in \inter_{j=0}^\infty X_i$; and}
\shortv{ $\sigma \in \inter_{j=0}^\infty X_i$ and,}
\fullv{\item} for each player $j$, each strategy $\sigma' \in X_j^k$ is
a best 
response to the beliefs $\mu_{\sigma',k}$. 
\fullv{\end{itemize}}
\edfn
Intuitively, $X_j^1$ consists of strategies that are best
responses to some belief of player $j$, and $X_j^{h+1}$ consists of
strategies 
in $X_j^h$ that are best responses to some belief of player $j$ with
support $X_{-j}^{h}$; that is, beliefs that assume that everyone else is
best reponding to some beliefs assuming that everyone else is responding
to some  beliefs assuming \ldots ($h$ times).

\pro\label{pro:rat} {\rm \cite{Pearce84}} A strategy is rationalizable
iff it is 
rationalizable$'$.  
\epro

We now give our epistemic characterizations of rationalizability.  
Let $\RAT$ be an abbreviation for $\RAT_1 \land \ldots \land \RAT_n$;
let $E \phi$ be an abbreviation of $B_1 \phi \land \ldots \land B_n
\phi$; and define $E^{k}
\phi$ for all $k$ inductively by taking $E^0 \phi$ to be $\phi$ and
$E^{k+1} \phi$ to be $E(E^k \phi)$.  
Common knowledge of $\phi$ holds iff
$E^k \phi$ holds for all $k \ge 0$.

We now give an epistemic characterization of rationalizability.  Part of
the characterization (the equivalence of (a) and (b) below)
is well known \cite{TW88}; it just says that a
strategy is rationalizable iff it can be played in a state where
rationality is common knowledge.
  
\thm\label{thm:charrat} The following are equivalent:
\begin{itemize}
\item[(a)] $\sigma$ is a rationalizable strategy for
$i$ in a game $\Gamma$; 
\item[(b)] there exists a measurable structure  $M$
that is appropriate for $\Gamma$ and a state $\omega$ such that
$\strat_i(\omega) = \sigma$ and $(M,\omega) \sat E^{k} \RAT$ for all
$k \ge 0$;  
\item[(c)]  there exists a measurable structure  $M$
that is appropriate for $\Gamma$ and a state $\omega$ such that
$\strat_i(\omega) = \sigma$ and $(M,\omega) \sat \dbi E^{k} \RAT$ for all
$k \ge 0$;  
\item[(d)]  there exists a structure  $M$
that is appropriate for $\Gamma$ and a state $\omega$ such that
$\strat_i(\omega) = \sigma$ and $(M,\omega) \sat \dbi E^{k} \RAT$ for all
$k \ge 0$.  
\end{itemize}
\ethm

\shortv{This proof, and all others, can be found in the full paper,
available at\\
http://www.cs.cornell.edu/home/halpern/papers/admissibility.pdf.} 
\fullv{
\prf Suppose that $\sigma$ is rationalizable.  Choose 
$\Z_j \subseteq \Sigma_j(\Gamma)$ and measures $\mu_{\sigma'}$ for each
strategy $\sigma' \in \Z_j$ guaranteed to exist by
Definition~\ref{rat1}.  Define an appropriate structure 
$M = (\Omega,\strat,\F, \PR_1, \ldots, \PR_n)$, where
\begin{itemize}
\item $\Omega = \Z_1 \times \cdots \times \Z_n$;
\item $\strat_i(\vec{\sigma}) = \sigma_i$;
\item $\F$ consist of all subsets of $\Omega$;
\item $\PR_i(\vec{\sigma})(\vec{\sigma}')$ is $0$ if $\sigma'_i \ne
\sigma_i$ and is $\mu_{\sigma_i}(\sigma'_{-i})$ otherwise.
\end{itemize}
Since each player is best responding to his beliefs at every state, it
is easy to see that $(M,\vec{\sigma}) \sat \RAT$ for all states
$\vec{\sigma}$.  It easily follows (formally, by 
induction on $k$), that $(M,\vec{\sigma}) \sat E^k \RAT$.
Clearly $M$ is measurable.
This shows that (a) implies (b).

The fact that (b) implies (c) is immediate, since if $E^{k+1} \phi$
logically implies $B_i E^k \phi$, which in turn logically implies
$\dbi_i E^k \phi$ for all $k$ and all formulas $\phi$.  The fact that
(c) implies (d) is also immediate.

Finally, to see that (d) implies (a), suppose that 
$M$ is a structure appropriate for $\Gamma$ and $\omega$ is a state in
$M$ such that $\strat_i(\omega) = \sigma$ and $(M,\omega) \sat \dbi
E^{k} \RAT$ for all $k \ge 0$.   For each player $j$, define the
formulas $C^k$ inductively by taking $C^0_j$ to be $\true$ and $C^{k+1}_j$
to be $\RAT_j \land B_j (\land_{j' \ne j}C^k_{j'})$.  
An easy induction
shows that for $k > 1$, $C^{k}_j$ is equivalent to $\RAT_j \land B_j
(E^0 \RAT \land \ldots \land E^{k-2} \RAT)$
in appropriate structures. 
Define $X^k_j = \{\strat_j(\omega'): (M,\omega') \sat C^k_j\}$.
If $\sigma' \in X^k_j$ for $k \ge 1$,
choose some state $\omega'$ such that $(M,\omega') \sat \RAT_j \land
B_jE^{k-2}\RAT$ and $\strat_j(\omega') = \sigma'$, and define  
$\mu_{\sigma',k}$ to be the projection of $\PR_j(\omega')$ onto
$\Sigma_{-j}$.  It easily follows that the support of $\mu_{\sigma',k}$
is $X^{k-1}_{-j}$ and that $\sigma'$ is a best response with respect to
$\mu_{\sigma,k}$.  Finally, since $(M,\omega) \sat \dbi E^k
\RAT$ for all $k \ge 0$, it easily follows that $\sigma =
\strat_i(\omega) \in \inter_{k=0}^\infty X_i^k$.  Thus, by
Definition~\ref{rat2}, $\sigma$ is rationalizable$'$ and, by
Proposition~\ref{pro:rat}, $\sigma$ is rationalizable.
\eprf
}

We now characterize iterated deletion of strongly dominated
(resp., weakly dominated) strategies.  

\dfn Strategy $\sigma$ for player is $i$ \emph{strongly dominated by 
$\sigma'$ with respect to $\Sigma'_{-i} \subseteq \Sigma_{-i}$} if 
$u_i(\sigma, \tau_{-i}) > u_i(\sigma, \tau_{-i})$ for all $\tau_{-i} \in
\Sigma'_{-i}$.  
Strategy $\sigma$ for player is $i$ \emph{weakly dominated by 
$\sigma'$ with respect to $\Sigma'_{-i} \subseteq \Sigma_{-i}$} if 
$u_i(\sigma, \tau_{-i}) \ge u_i(\sigma, \tau_{-i})$ for all $\tau_{-i} \in
\Sigma'_{-i}$ and   $u_i(\sigma, \tau'_{-i}) > u_i(\sigma, \tau'_{-i})$
for some $\tau'_{-i} \in \Sigma'_{-i}$.

Strategy $\sigma$ for player $i$ survives $k$ rounds of iterated
deletion of strongly dominated (resp., weakly dominated) strategies if, 
for each player $j$, there exists a
sequence $X_j^0, X_j^1, X_j^2, \ldots, X_j^k$ of sets of strategies for
player 
$j$ such that $X_j^0 = \Sigma_j$ and, if $h < k$, then $X_j^{h+1}$
consists of the strategies in $X_j^h$ not strongly (resp., weakly)
dominated by any strategy with respect to $X_{-j}^h$, and $\sigma \in
X_i^k$.  Strategy $\sigma$ survives iterated deletion of strongly
dominated (resp., weakly dominated) strategies if it survives $k$ rounds
of iterated deletion for all $k$. 
\edfn

The following well-known result 
connects strong and weak dominance to best responses.
\pro\label{pro:Pearce} {\rm \cite{Pearce84}}
\begin{itemize}
\item A strategy $\sigma$ for player $i$ is not strongly dominated by
any strategy with
respect to $\Sigma'_{-i}$ iff there is a belief
$\mu_\sigma$ of player $i$ whose support is a subset of $\Sigma'_{-i}$
such that $\sigma$ is a best response with respect to $\mu_\sigma$.  
\item A strategy $\sigma$ for player $i$ is not weakly dominated by any
strategy with respect to $\Sigma'_{-i}$ iff there 
is a belief $\mu_\sigma$ of player $i$ whose support is all of
$\Sigma'_{-i}$ such that $\sigma$ is a best response with respect to
$\mu_\sigma$.   
\end{itemize}
\epro

It immediately follows from 
Propositions~\ref{pro:rat} and~~\ref{pro:Pearce} (and is well known)
that a strategy is 
rationalizable iff it survives 
iterated deletion of strongly dominated strategies.  Thus, the
characterization of rationalizability in Theorem~\ref{thm:charrat} is
also a characterization of strategies that survive iterated deletion of
strongly dominated strategies.  
To characterize iterated deletion of weakly dominated strategies, we
need to enrich the langauge $\L_1$ somewhat.  Let
$\L_2(\Gamma)$ be 
the extension of $\L_1$ that includes a 
primitive proposition $\play_i(\sigma)$ for each player $i$ and strategy
$\sigma \in \S_i$, and is also closed off under the modal operator 
$\Diamond$.
We omit the parenthetical $\Gamma$ when it is clear from context.
We extend the truth relation 
to $\L_2$ in probability structures appropriate for $\Gamma$ as
follows: 
\begin{itemize}
\item $(M,\omega) \sat \play_i(\sigma)$ iff $\omega \in
\intension{\sigma}$.  
\item $(M,\omega) \sat \Diamond \phi$ iff there is some structure $M'$
appropriate for $\Gamma$ and state $\omega'$ such that $(M',\omega')
\sat \phi$.
\end{itemize}
Intuitively, $\Diamond \phi$ is true if there is some state and
structure where $\phi$ is true; 
that is, if $\phi_i$ is satisfiable.  Note that if $\Diamond \phi$ is
true at some state, then it is true at all states in all structures.

Let $\play (\vec{\sigma})$ be an abbreviation for 
$\land_{j=1\,}^n \play_j(\sigma_j)$, and 
let $\play_{-i}(\sigma_{-i})$ be an abbrevation for $\land_{j\ne i}
\play_j(\sigma_j)$.  Intuitively, $(M,\omega) \sat \play(\vec{\sigma})$
iff $\strat(\omega) = \sigma$, and $(M,\omega) \sat
\play_{-i}(\sigma_{-i})$ if, at $\omega$,  
the players other than $i$ are playing strategy profile $\sigma_{-i}$.
Define the formulas $D^k_j$ inductively by taking $D^0_j$ 
to be the formula $\true$,  and $D^{k+1}_j$ to 
be
an abbreviation of 
$$\RAT_j \land B_j (\land_{j' \ne j}D^k_{j'}) \land (\land_{\sigma_{-j}
\in \S_{-j}} \Diamond(\play_{-j}(\sigma_{-j}) \land (\land_{j' \ne j}
D^k_{j'})) \rimp \dbj (\play_{-j}(\sigma_{-j})).$$ 

\fullv{It is easy to see that $D^k_j$ implies the formula $C^k_j$ defined in
the proof of Theorem~\ref{thm:charrat}, and hence implies}
\shortv{It easily follows by induction that $D^k_j$ implies} 
 $\RAT_j \land B_j
(E^0 \RAT \land \ldots \land E^{k-2} \RAT)$. 
But $D^k_j$ requires more;
it requires  that player $j$ assign positive
probability to each strategy profile for the other players that is
compatible  
with $D^{k-1}_{-j}$.  

\thm\label{thm:charwd} The following are equivalent:
\begin{itemize}
\item[(a)] the strategy $\sigma$ for player $i$ survives $k$ rounds of
iterated  
deletion of weakly dominated strategies;
\item[(b)] for all $k' \le k$, there is a measurable structure  $M^{k'}$
appropriate for $\Gamma$ and a state $\omega^{k'}$ in $M^{k'}$ such that
$\strat_i(\omega^{k'}) = \sigma$ and $(M^{k'},\omega^{k'}) \sat D^{k'}_i$;  
\item[(c)] for all $k' \le k$, there is a structure  $M^{k'}$
appropriate for $\Gamma$ and a state $\omega^{k'}$ in $M^{k'}$ such that
$\strat_i(\omega^{k'}) = \sigma$ and $(M^{k'},\omega^{k'}) \sat D^{k'}_i$.  
\end{itemize}
%
\iffalse
In addition,  there is a finite structure $M^k =
(\Omega^k,\strat,\F, \PR_1, \ldots, \PR_n)$ such that 
$\Omega^k = \{(k',\vec{\sigma}): k' \le k, \vec{\sigma} \in X^{k'}_1
\times \cdots \times X^{k'}_n\}$, $\strat(k',\vec{\sigma}) =
\vec{\sigma}$, $\F = 2^{\Omega^k}$, where $X^{k'}_j$ consists of all
strategies for player $j$ thaat survive $k$ rounds of iterated deletion
of weakly dominated strategies, 
and, for all states 
$(k', \vec{\sigma}) \in \Omega^k$,  
$(M^k,(k',\vec{\sigma})) \sat D_1^{k'} \land \ldots \land D_n^{k'}$.
\fi
%
%
In addition,  there is a finite structure $\overline{M}^k =
(\Omega^k,\strat,\F, \PR_1, \ldots, \PR_n)$ such that 
$\Omega^k = \{(k',i,\vec{\sigma}): k' \le k, 1 \le i \le n, \vec{\sigma}
\in X^{k'}_1 \times \cdots \times X^{k'}_n\}$, $\strat(k',i,\vec{\sigma}) =
\vec{\sigma}$, $\F = 2^{\Omega^k}$, where $X^{k'}_j$ consists of all
strategies for player $j$ that survive $k'$ rounds of iterated deletion
of weakly dominated strategies
and, for all states 
$(k', i, \vec{\sigma}) \in \Omega^k$,  
$(\overline{M}^k,(k',i, \vec{\sigma})) \sat \land_{j \ne i} D_j^{k'}$.
\ethm

\fullv{
\prf 
We proceed by induction on $k$, proving both the equivalence of (a),
(b), and (c) 
and the existence of a structure $\overline{M}^k$ with the required properties.

The result clearly holds if $k=0$.  
Suppose that the result holds for $k$; we show that it holds for $k+1$.
We first show that (c) implies (a).  Suppose that 
 $(M^{k'},\omega^{k'}) \sat D^{k'}_j$ and $\strat_j(\omega^{k'}) =
\sigma_j$ for all $k' \le k+1$.   
It follows that $\sigma_j$ is a best response
to the belief $\mu_{\sigma_j}$ on the strategies
of other players induced by $\PR^{k+1}_j(\omega)$.  Since 
$(M^{k+1},\omega^{k+1}) \sat B_j(\land_{j' \ne j} D^{k}_{j'})$, it
follows from the 
induction hypothesis that 
the support of $\mu_{\sigma_j}$ is contained in $X^k_{-j}$.  Since
$(M,\omega) \sat \land_{\sigma_{-j} \in \S_{-j}}
(\Diamond(\play_{-j}(\sigma_{-j}) \land (\land_{j\ne i}
D^k_j)) \rimp 
\dbj (\play_{-j}(\sigma_{-j})))$, it follows from the
induction hypothesis that the support of $\mu_{\sigma_j}$ is all of 
$X^k_{-j}$.  Since $(M^{k'},\omega^{k'}) \sat D^{k'}_j$ for $k' \le k$,
it follows from the 
induction hypothesis that $\sigma_j
\in X^k_j$.  Thus, $\sigma_j \in X^{k+1}_j$.  

We next construct the structure $\overline{M}^{k+1} =
(\Omega^{k+1},\strat,\F, \PR_1, \ldots, \PR_n)$.  
As required, we define
$\Omega^{k+1} = \{(k',i,\vec{\sigma}): k' \le k+1, 1 \le i \le n,
\vec{\sigma} \in
X^{k'}_1 \times \cdots 
\times X^{k'}_n\}$, $\strat(k', i, \vec{\sigma}) = \vec{\sigma}$, $\F =
2^{\Omega^{k+1}}$.  
For a state $\omega$ of the form $(k', i,\vec{\sigma})$,
since $\sigma_j \in X^{k'}_j$,
by Proposition \ref{pro:Pearce},
there exists a distribution
$\mu_{k',\sigma_j}$ whose support is all of $X^{k-1}_{-j}$ such that
$\sigma_j$ is a best response to $\mu_{\sigma_j}$.  
Extend
$\mu_{k',\sigma_j}$ to a distribution $\mu_{k',i,\sigma_j}'$ on $\Omega^{k+1}$
as follows:
\begin{itemize}
\item for $i\neq j$, let $\mu_{k',i,\sigma_j}'(k'',i',\vec{\tau}) = 
\mu_{k',\sigma_j}(\vec{\tau}_{-j})$ if $i'=j, k'' = k'-1$, and $\tau_j =
\sigma_j$, and 0 otherwise;   
\item$\mu_{k',j,\sigma_j}'(k'',i',\vec{\tau}) = 
\mu_{k',\sigma_j}(\vec{\tau}_{-j})$ if $i'=j, k'' = k'$, and $\tau_j = \sigma_j$, and 0 otherwise.  
\end{itemize}
Let $\PR_j(k',i,\vec{\sigma}) =
\sigma'_{k',i,\sigma_j}$.  We leave it to the reader to check that this
structure is appropriate.  
An easy
induction on $k'$ now shows that 
$(\overline{M}^{k+1},(k',i,\vec{\sigma})) \sat \land_{j \ne i} D^{k'}_{j}$ for 
$i = 1, \ldots, n$.

To see that (a) implies (b), suppose that $\sigma_j \in X^{k+1}_j$.
Choose a state $\omega$ in 
$\overline{M}^{k+1}$ of the form $(k+1,i,\vec{\sigma})$, where $i \neq j$.  
As we just showed,
$(\overline{M}^{k+1}, (k',i, \vec{\sigma}) \sat D^{k'}_j$, and
$\strat_j(k',i, \vec{\sigma}) = \sigma_j$.  Moreover, $\overline{M}^{k+1}$ is
measurable (since $\F$ consists of all subsets of $\Omega^{k+1}$).

Clearly (b) implies (c).
\eprf
}

%
\iffalse
We remark that, although in appropriate structures, an agent knows his
strategy, he does not know his beliefs.  That is, he may assign positive
probability to states where he has different beliefs.  We use this fact
in the construction of $M^{k+1}$.  The support of player $i$'s belief
at a state $(k+1,\vec{\sigma})$ consists states of the form
$(k,\vec{tau})$, but at these states, player $i$'s beliefs have support
consisting of states of the form $(k-1,\vec{\tau})$.  If we want to
consider only structures where player $i$ knows his beliefs, we must
modify the $D^k_j$ slightly, so that $D^{k+1}_j$ is an abbreviation
%
for
%
%
%
%
%
$$\RAT_j \land B_j (\land_{j' \ne j}D^k_{j'}) \land (\land_{\sigma_{-j}
\in \S_{-j}} \Diamond(\play_{-j}(\sigma_{-j}) \land (\land_{j' \ne j}
D^k_{j'})) \rimp \dbj (\play_{-j}(\sigma_{-j}) \land (\land_{j' \ne j}
D^k_{j'})).$$ 
%
%
%
%
%
With this modification, we can
restrict attention to structures where player $i$ knows his beliefs,
essentially the same proof.
\fi

\cor\label{cor:wd}  The following are equivalent:
\begin{itemize}
\item[(a)] the strategy $\sigma$ for player $i$ survives iterated 
deletion of weakly dominated strategies;
\item[(b)] there is a measurable structure  $M$
that is appropriate for $\Gamma$ and a state $\omega$ such that
$\strat_i(\omega) = \sigma$ and $(M,\omega) \sat \dbi D^k_i$ for all $k
\ge 0$;  
\item[(c)] there is a structure  $M$
that is appropriate for $\Gamma$ and a state $\omega$ such that
$\strat_i(\omega) = \sigma$ and $(M,\omega) \sat \dbi D^k_i$ for all $k
\ge 0$.  
\end{itemize}
\ecor

\commentout{
\prf Clearly if (c) holds, for all $k$, there is a structure $M$ and
state $\omega_k$ such that $\strat_i(\omega_k) = \sigma$ such that
$(M,\omega_k) \sat D^k_i$.  Thus, by Theorem~\ref{thm:charwd}, $\sigma$
survives $k$ rounds of iterated deletion for all $k$, and hence survies
iterated deletion.  

To see that (a) implies (b), consider the structures $M^k$ constructed
in Theorem~\ref{thm:charwd}.  There must exist a state 
$\omega_k \in \Omega^k$ such that 
$\strat^k_i(\omega_k) = \sigma$, 
and $(M^k,\omega_k) \sat D^k_i$.   We construct a new structure
$M^\infty$ that results from ``pasting together'' $M^k$, $k = 0, 1, 2, 
\ldots$,
such that $(M^\infty,\omega) \sat  \dbi D^k_i$ for all $k \ge 0$.
The construction is formalized in Lemma~\ref{lem:paste}.

Clearly (b) implies (c).  
\eprf
}

Note that there is no analogue of Theorem~\ref{thm:charrat}(b) here.
This is because there is no state where $D^k_i$ holds for all $k \ge 0$;
it cannot be the case that $i$ places positive probability on all
strategies (as required by $D^k_1$) and that $i$ places positive
probability only on strategies that survive one round of iterated
deletion (as required by $D^k_2$), unless all strategies survive one
round on iterated deletion.  We can say something slightly weaker
though.  There is some $k$ such that the process of iterated deletion
converges; that is, $X^k_j = X^{k+1}_j$ for all $j$ (and hence
$X^k_j = X^{k'}_j$ for all $k' \ge k$).  That means that there is a
state where $D^{k'}_i$ holds for all $k' > k$.  Thus, we can show that 
a strategy $\sigma$ for player $i$ survives iterated deletion of weakly
dominated strategies iff there exists a $k$ and a state $\omega$ such
that $\strat_i(\omega) = \sigma$ and $(M,\omega) \sat D^{k'}_i$ for all
$k' > k$.  Since $C^{k+1}_i$ implies $C^k_i$, an anlagous results holds
for iterated deletion of strongly dominated strategies, with $D^{k'}_i$
replaced by $C^{k'}_i$.

It is also worth noting that in a state  where $D^k$ holds, an agent
does \emph{not} 
consider all strategies possible, but only the ones consistent with the
appropriate level of rationality.  We could require the agent to
consider all strategies possible by using LPS's or nonstandard
probability.  The only change that this would make to our
characterization is that, if we are using nonstandard probability, we
would interpret $B_i \phi$ to mean that 
$\phi$ holds with probability infinitesimally close to 1, while $\dbi
\phi$ would mean that $\phi$ holds with probability whose standard part
is positive (i.e., non-infinitesimal probability).  We do not pursue
this point further.

\section{Strong Admissibility}\label{sec:strong}

We have formalized iterated admissibility by saying that an agent
consider possible all strategies consistent with the appropriate
rationality assumption.  But why focus just on strategies?  We now consider a
stronger admissibility requirement that we call, not surprisingly,
\emph{strong admissibility}.  Here we require, intuitively, that
\emph{all} an agent knows about the other agents is that they satisfy
the appropriate rationality assumptions.  Thus, the agent ascribes
positive probability to all beliefs that the other agents could have as
well as all the strategies they could be using.  By considering strong
admissibility, we will be able to relate work on ``all I know''
\cite{HalLak94,Lev5}, BFK's notion of complete structures, and admisibility.

Roughly speaking, we interpret ``all agent $i$ knows is $\phi$'' as
meaning that agent $i$ believes $\phi$, and considers possible every
formula about the other players' strategies and beliefs consistent with
$\phi$.  Thus, what
``all I know'' means is very sensitive to the choice of language.  
Let $\L^0$ be the language whose only formulas are (Boolean combinations
of) formulas of the form $\play_i(\sigma)$, $i = 1, \ldots, n$, $\sigma
\in \Sigma_i$.  Let $\L^0_i$ consist of just the formulas of the form
$\play_i(\sigma)$, and let $\L^0_{-i} = \union_{j \ne i} \L^0_j$. 
Define $O^-_i\phi$ to be
an abbreviation for $B_i \phi \land (\land_{\psi \in \L^0_{-i}} \Diamond
(\phi \land \psi) \rimp \dbi \psi$).
Then it is easy to see that $D^{k+1}_j$ is just $\RAT_j \land O^-_j
(\land_{j' \ne j} D^k_{j'})$.  

We can think of $O^-_i \phi$ as saying ``all agent $i$ knows with
respect to the language $\L^0$ is $\phi$.''
The language $\L^0$ is quite weak.  To relate our results to those of
BFK,  even the language $\L^2$ is too weak, since it does not
allow an agent to express probabilistic beliefs.
Let $\L^3(\Gamma)$ be the language that extends $\L^2(\Gamma)$ by allowing
formulas of the form $\pr_i(\phi) \ge \alpha$ and $\pr_i(\phi) >
\alpha$, where $\alpha$ is a rational number in $[0,1]$;
$\pr_i(\phi) \ge \alpha$ can be read as ``the probability of $\phi$
according to $i$ is at least $\alpha$'', and similarly for $\pr_i(\phi) >
\alpha$.  We allow
nesting here, so that we can have a formula of the form
$\pr_j(\play_i(\sigma) \land \pr_k(\play_i(\sigma')) > 1/3) \ge 1/4$.  
As we would expect,
\begin{itemize}
\item $(M,\omega) \sat \pr_i(\phi)$ iff $\PR_i(\omega)(\intension{\phi})
\ge \alpha$.
\end{itemize}
The restriction to $\alpha$ being rational allows the language to be
countable.   
However, as we now show, it is not too serious a 
restriction.  

Let $\L^4(\Gamma)$ be the language that extends $\L^2(\Gamma)$ by closing off 
under countable conjunctions, so that if $\phi_1, \phi_2, \ldots$ are
formulas, then so is $\land_{m=1}^\infty \phi_m$, and formulas of the form
$\pr_i(\phi) > \alpha$, where $\alpha$ is a real number in $[0,1]$.
(We can express $\pr_i(\phi) \ge \alpha$ as the countable 
conjunction $\land_{\beta < \alpha, \beta \in Q \inter [0,1]}
\pr_i(\phi) > \beta$, where $Q$ is the set of rational numbers, so there
is no need to include formulas of the form $\pr_i(\phi) \ge \alpha$
explicitly in $\L^4(\Gamma)$.)   We omit the parenthetical $\Gamma$ in
$\L^3(\Gamma)$ and $\L^4(\Gamma)$ when the game $\Gamma$ is clear from
context.  
A subset $\Phi$ of $\L^3$ is
$\L^3$-\emph{realizable} if there 
exists an appropriate structure $M$ for $\Gamma$ and state $\omega$ in
$M$ such that, for all formulas $\phi \in \L^3$,  $(M,\omega) \sat 
\phi$ iff $\phi \in \Phi$.%
\footnote{For readers familiar with standard completeness proofs in
modal logic, if we had axiomatized the logic we are implicitly using
here, the $\L^3$-realizable sets would just be the maximal consistent
sets in the logic.}
We can similarly define what it means for a subset of $\L^4$ to be
$\L^4$-realizable.

\lem\label{lem:realizable} Every $\L^3$-realizable set can be uniquely
extended to an $\L^4$-realizable set. \elem

\fullv{
\prf It is easy to see that every $\L^3$-realizable set can be extended 
to an $\L^4$-realizable set.  For suppose that $\Phi$ is
$\L^3$-realizable.  Then there is some state $\omega$ and structure $M$
such that, for every formula $\phi \in \L^3$, we have that $(M,\omega)
\sat \phi$ iff $\phi \in \Phi$.   Let $\Phi'$ consist of the $\L^4$
formulas true at $\omega$.  Then clearly $\Phi'$ is an $\L^4$-realizable
set that extends $\Phi$.

To show that the extension is unique, suppose that there are two
$\L^4$-realizable sets, say $\Phi_1$ and $\Phi_2$, that extend $\Phi$.
We want to show that $\Phi_1 = \Phi_2$.  To do this, we consider two
language, $\L^5$ and $\L^6$, intermediate between $\L^3$ and $\L^4$.  

Let $\L^5$ be the language that extends $\L^2$ by 
closing off 
under countable conjunctions and formulas of the form 
$\pr_i(\phi) > \alpha$, where $\alpha$ is a rational number in $[0,1]$.
Thus, in $\L^5$, we have countable conjunctions and disjunctions, but
can talk explicitly only about rational probabilities.  Nevertheless,
it is easy to see that for every formula $\phi \in \L^4$, there
is an formula equivalent formula $\phi' \in \L^5$, since if
$\alpha$ is a real number, then $\pr_i(\phi) > \alpha$ is equivalent   
to $\lor_{\beta > \alpha, \, \beta \in [0,1]\inter Q\,} \pr_i(\phi) > \beta$
(an infinite disjunction $\lor_{i=1}^\infty \phi_i$ can be viewed as an
abbreviation for $\neg  \land_{i=1}^\infty \neg \phi_i$).

Next, let $\L^6$ be the result of closing off formulas in
$\L^3$ under countable conjunction and disjunction.  Thus, in
$\L^6$, we can apply countable conjunction and disjunction only at the
outermost level, not inside the scope of $\pr_i$.  
We claim that for every formula $\phi \in \L^5$, there is an
equivalent formula in $\L^6$.  More precisely, for every formula
$\phi \in \L^5$, there exist
formulas $\phi_{ij} \in \L^3$, $1 \le i,j  < \infty$ such that 
$\phi$ is equivalent to $\land_{m=1}^\infty \lor_{n=1}^\infty
\phi_{mn}$.   We prove this by induction on the structure of $\phi$.
If $\phi$ is $\RAT_i$, $\play_i(\sigma)$, or $\true$, then the statement
is clearly true.  The result is immediate from the induction hypothesis
if $\phi$ is a countable conjunction.  If $\phi$ has the form $\neg
\phi'$, we apply the induction hypothesis, and observe 
that $\neg (\land_{m=1}^\infty \lor_{n=1}^\infty \phi_{mn})$ is
equivalent to $\lor_{m=1}^\infty \land_{n=1}^\infty \neg \phi_{mn}$.  
We can convert this to a conjunction of disjunctions by distributing the
disjunctions over the conjunctions in the standard way (just as 
$(E_1 \inter E_2) \union (E_3 \inter E_4)$ is equivalent to
$(E_1 \union E_3) \inter (E_1 \union E_4) \inter (E_2 \union E_3) \inter
(E_2 \union E_4)$).  Finally, if $\phi$ has the form $\pr_i(\phi') > 
\alpha$, we apply the induction hypothesis, and observe that 
$\pr_i(\land_{m=1}^\infty \lor_{n=1}^\infty \phi_{mn}) > \alpha$ is
equivalent to $$\lor_{\alpha' > \alpha, \alpha' \in Q \inter [0,1]}
\land_{M=1}^\infty \lor_{N=1}^\infty \pr_i(\land_{m=1}^M \lor_{n=1}^N
\phi_{mn}) > \alpha'.$$

The desired result follows, since if two states agree on all formulas in
$\L^3$, they must agree on all formulas in $\L^6$, and hence on all
formulas in $\L^5$ and $\L^4$. \eprf

The choice of language turns out to be significant for a
number of our results; we return to this issue at various points below.
}

With this background, we can define strong admissibility.  
Let $\L^3_i$ consist of all formulas in
$\L^3$ of the form $\pr_i(\phi) \ge \alpha$ and $\pr_i(\phi) >
\alpha$ ($\phi$ can mention $\pr_i$; it is only the outermost
modal operator that must be $i$).  Intuitively, $\L^3_i$
consists of the formulas describing $i$'s beliefs.  Let 
$\L^3_{i+}$ consist of $\L^3_i$ together with formulas of the form
$\true$, $\RAT_i$, and $\play_i(\sigma)$, for $\sigma \in\S_i$.  
Let $\L^3_{(-i)+}$ be an abbreviation for $\union_{j \ne i} \L^3_{j+}$.
We can similarly define $\L^4_i$ and $\L^4_{i+}$.

If $\phi \in \L^3_{(-i)+}$, define $O_i \phi$, read ``all agent $i$
knows (with respect to $\L^3$) is $\phi$,'' as an abbreviation for the
$\L^4$ formula  
$$B_i \phi \land (\land_{\psi \in
\L^3_{(-i)+}}  \Diamond(\phi \land \psi) \rimp \dbj \psi).$$  
Thus, $O_i \phi$ holds if agent $i$ believes $\phi$ but does not know
anything beyond that; he ascribes positive probability to all formulas
in $\L^3_{(-i)+}$ consistent with $\phi$.  This is very much in the
spirit of the Halpern-Lakemeyer \citeyear{HalLak94} definition of $O_i$
in the context of epistemic logic.

Of course, we could go further and define a notion of ``all $i$ knows''
for the language $\L^4$.  
Doing this would give a definition that is 
even closer to that of Halpern and Lakemeyer.
Unfortunately, we cannot require than 
agent $i$  ascribe positive probability to all the formulas  in
$\L^4_{(-i)+}$ consistent with $\phi$; in general, there will be
an uncountable number of distinct and mutually exclusive formulas
consistent with $\phi$, so they
cannot all be assigned positive probability.  This problem does not
arise with $\L^3$, since it is a countable language.  
Halpern and Lakemeyer could allow an agent to consider an uncountable
set of worlds possible, since they were not dealing with probabilistic
systems.  
This stresses the point that the notion of ``all I know'' is quite
sensitive to the choice of language.

Define the formulas $F^k_i$ inductively by taking $F^0_i$ 
to be the formula $\true$,  and $F^{k+1}_i$ to 
an abbreviation of 
$\RAT_i \land O_i (\land_{j \ne i}F^k_{j})$. 
Thus, $F^{k+1}_j$ says that $i$ is rational, believes that all the other
players satisfy level-$k$ rationality (i.e., $F^k_{j}$), and that is all
that $i$ knows.  An easy induction shows that $F^{k+1}_j$ implies that
$j$ is rational and $j$ 
believes  
that everyone believes ($k$ times) that everyone is
rational.  Moreover, it is easy to see that $F^{k+1}_j$ implies
$D^{k+1}_j$.   The difference is that instead of requiring just that $j$
assign positive probability to all strategy profiles compatible with
$F^{k}_{-j}$, it requires that $j$ assign positive probability to all
formulas compatible with $F^{k}_{-j}$.  
A strategy $\sigma_i$ for player $i$ is
\emph{$k$th-level strongly admissible} if it is consistent with $F^k_i$;
that is, if $\play_i(\sigma_i) \land F^k_i$ is satisfied in some state.
The next result shows that strong admissibility characterizes iterated
deletion, just as admissibility does.

\thm\label{thm:charwd1} The following are equivalent:
\begin{itemize}
\item[(a)] the strategy $\sigma$ for player $i$ survives $k$ rounds of
iterated  
deletion of weakly dominated strategies;
\item[(b)] for all $k' \le k$, there is a measurable structure  $M^{k'}$
appropriate for $\Gamma$ and a state $\omega^{k'}$ in $M^{k'}$ such that
$\strat_i(\omega^{k'}) = \sigma$ and $(M^{k'},\omega^{k'}) \sat F^{k'}_i$;
\item[(c)] for all $k' \le k$, there is a structure  $M^{k'}$
appropriate for $\Gamma$ and a state $\omega^{k'}$ in $M^{k'}$ such that
$\strat_i(\omega^{k'}) = \sigma$ and $(M^{k'},\omega^{k'}) \sat F^{k'}_i$;
\end{itemize}
\ethm

\fullv{
\prf The proof is similar in
spirit to the proof of Theorem~\ref{thm:charwd}.  We again proceed by
induction on $k$.  The result clearly holds for $k=0$.  If $k=1$, the
proof that (c) implies (a) is essentially identical to that of
Theorem~\ref{thm:charwd}; we do not repeat it here.  

To prove that (a)
implies (b), we need the
following three lemmas; the first shows that a formula is always
satisfied in a state that has probability 0; the 
second shows that  that we can get a
new structure with a world where agent $i$ ascribes positive probability
to each of a countable collection of satisfiable formulas in
$\L_{-i}^3$; and the third shows that formulas in 
$\L^4_{i+}$ for different players $i$ are independent; that is, if 
$\phi_i \in \L^4_{i+}$ is satisfiable, then so is 
$\phi_1 \land \ldots \land \phi_n$.  

\lem\label{lem:paste0} If $\phi \in \L^4$ is satisfiable in a
measurable structure, then there
exists a measurable structure $M$ and state $\omega$ such that
$(M,\omega)\sat \phi$, $\{\omega\}$ is measurable,
$\PR_j(\omega)(\{\omega\}) = 0$ for $j = 1,\ldots,n$.  \elem

\prf Suppose that
$(M',\omega') \sat\phi$, where $M' = (\Omega',\strat',\F', \PR_1', \ldots,
\PR_n')$.
Let $\Omega = \Omega' \union \{\omega\}$, where 
where $\omega$ is a fresh state; let $\F$ be the smallest
$\sigma$-algebra that contains $\F$ and $\{\omega\}$;
let $\strat$ and $\PR_j$ agree with
$\strat'$ and $\PR_j'$ when restricted to states in $\Omega'$ (more
precisely, if $\omega'' \in \Omega'$, then 
$\PR_j(\omega'')(A) = \PR_j'(\omega'')(A \inter \Omega')$ for $j =
1,\ldots, n$).    
Finally, define $\strat_i(\omega)= \strat_i(\omega')$, and take
$\PR_j(\omega)(A) = \PR_j'(\omega')(A \inter \Omega')$ for $j =
1,\ldots,n$.   Clearly $\{\omega\}$ is measurable, and
$\PR_j(\omega)(\{\omega\}) = 0$ for $j = 1,\ldots,n$.  An easy induction
on structure shows that for all formulas $\psi$,
(a) $(M,\omega) \sat \psi$ iff  $(M,\omega') \sat \psi$, and 
(b) for all states $\omega'' \in \Omega'$, we have that 
$(M,\omega'') \sat \psi$ iff $(M',\omega'') \sat \psi$.
It follows that $(M,\omega) \sat \phi$, and that $M$ is measurable.
\eprf
 
\lem\label{lem:paste} Suppose that $\vec{\sigma} \in \vec{\Sigma}$,
$\Phi'$ is a countable collection of formulas in $\L^4_{-i}$, $\phi \in
\L^4_{-i}$, 
and $\Sigma'_{-i}$ is a set of strategy profiles in $\Sigma_{-i}$ such
that (a) for each formula $\phi' \in \Phi'$, there exists some profile
$\sigma_{-i} \in \Sigma'_{-i}$ such that 
$\phi \land \phi' \land \play_{-i}(\sigma_{-i})$ is satisfied in a
measurable structure, and
(b) for each profile $\sigma_{-i} \in \Sigma'_{-i}$,
$\play_{-i}(\sigma_{-i})$ is one of the formulas in $\Phi'$.
Then there exists a measurable structure $M$ and state
$\omega$ such that $\strat(\omega) = \vec{\sigma}$, $(M,\omega) \sat
\play_{-j}(\sigma_{-i}) \ge \alpha$ 
iff $\mu_j(\sigma_{-i}) \ge \alpha$ (that is, $\mu_{-i}$ agrees with
$\PR_i(\omega)$ when marginalized to strategy profiles in
$\Sigma'_{-i}$), and $(M,\omega) \sat B_i \phi \land \dbi \phi'$  for
all $\phi' \in \Phi'$.
\elem

\prf Let $\Phi'$ and $\Sigma'_{-i}$ be as in the
statement of the lemma.  Suppose that $\Phi' = \{\phi_1, \phi_2, \ldots,
\ldots\}$.  By assumption, for each formula $\phi_k \in \Phi'$, there
exists some strategy profile $\sigma'_{-i} \in \Sigma'_{-i}$, 
measurable structure
$M^k = (\Omega^k, \strat^k, \F^k, \PR_1^k, \ldots, \PR_n^k)$, and
$\omega^k \in \Omega^k$ such that 
$(M^k,\omega^k) \sat \phi \land \phi_k \land \play_{-i}(\sigma_{-i}')$,
for $k = 1, 2, \ldots$.
By Lemma~\ref{lem:paste0}, we can assume without
loss of generality that $\{\omega^k\} \in \F^k$ and
$\PR_j^k(\omega^k)(\{\omega^i\})  = 
0$.  Define $M^\infty = (\Omega^\infty, \strat^\infty, \F^\infty,
\PR_1^\infty, \ldots, \PR_n^\infty)$ as follows:
\begin{itemize}
\item $\Omega^\infty = \union_{k=0}^\infty \Omega^k \union
\{\omega\}$, where $\omega$ is a fresh state; 
\item $\F^\infty$ is the
smallest  $\sigma$-algebra that contains $\{\omega\} \union \F_1 \union
\F_2 \union \ldots$; 
\item $\strat^\infty$ agrees with $\PR_j^k$ when restricted to states in
$\Omega^k$, except that $\strat^\infty_i(\omega^k) = \sigma_i$ and
$\strat^\infty(\omega) = \vec{\sigma}$;
\item $\PR_j^\infty$ agrees with
$\PR_j^k$ when restricted to states in $\Omega^k$ (more
precisely, if $\omega' \in \Omega^k$, then 
$\PR_j^\infty(\omega')(A) = \PR_j^k(\omega')(A \inter \Omega^k)$, except
that 
$\PR_i^\infty(\omega) = \PR_i^\infty(\omega^1) = \PR_i(\omega^2) =
\cdots$ is defined to be a distribution with support $\{\omega^1,
\omega^2, \ldots\}$ (so that all these states are given positive
probability) such that $\PR_i^\infty(\omega)$ agrees $\mu$ when
marginalized to profiles in $\Sigma_{-i}$, 
and $\PR_j^\infty(\omega)(\{\omega\}) = 1$ for $j \ne i$.
It is easy to see that our
assumptions guarantee that this can be done.  
\end{itemize}

We can now prove by a straightforward induction on the structure of
$\psi$ that (a) for 
all formulas $\psi$, $k = 1, 2, 3, \ldots$, and 
states $\omega' \in \Omega^k - \{\omega^k\}$, we have that 
$(M^k,\omega') \sat \psi$ iff $(M^\infty,\omega') \sat \psi$; and 
(b)  for all formulas $\psi \in \L^4_{(-i)^+}$, $k = 1, 2, 3, \ldots$, and
$(M^k,\omega^k) \sat \psi$ iff $(M^\infty,\omega^k) \sat \psi$. (Here it
is important that $\PR_j^\infty(\omega^k) = \PR_j^k(\omega) = 0$ for $j
\ne i$; this ensures that $j$'s beliefs about $i$'s strategies and beliefs
unaffected by the fact that  $\strat_i^k(\omega^k) \ne
\strat_i^\infty(\omega^k)$ and $\PR_i^k(\omega^k) \ne
\PR_i^\infty(\omega^k)$.)  It easily follows that 
$(M^\infty,\omega) \sat B_i \phi \land \dbi \phi'$ for all $\phi'
\in \Phi'$. \eprf

\lem\label{lem:paste1} If $\phi_i \in \L^4_{i+}$ is
satisfiable for $i = 1, \ldots, n$, 
then $\phi_1 \land \ldots \land \phi_n$ is satisfiable.
\elem

\prf Suppose that $(M^i,\omega^i) \sat \phi_i$, where 
$M^i = (\Omega^i,\strat^i,\F^i,\PR_1^i,\ldots, \PR_n^i)$ and $\phi_i \in
\L^4_{i+}$.  By Lemma~\ref{lem:paste0}, we again assume without loss of
generality that $\{\omega^i\} \in \F^i$ and
$\PR_j(\omega^i)(\{\omega^i\}) = 0$.  Let $M^* =
(\Omega^*,\strat^*,\F^*, \PR_1^*, \ldots, \PR_n^*)$, where 
\begin{itemize}
\item $\Omega^* = \union_{i=1}^n \Omega^i$;
\item $\F^*$ is the smallest $\sigma$-algebra containing $\F^1 \union
\ldots \union \F^n$;
\item $\strat^*$ agrees with $\strat^j$ on states in $\Omega^j$ except
that $\strat^*_i(\omega^j) = \strat^i_i(\omega^i)$ (so that
$\strat^*(\omega^1) = \cdots = \strat^*(\omega^n)$);
\item $\PR^*_i$ agrees with $\PR^j_i$ on states in $\Omega^j$ except
that $\PR^*_i(\omega^j) = \PR^i_i(\omega^i)$ (so that
$\PR_i^*(\omega^1) = \cdots = \PR_i^*(\omega^n) = \PR_i^i(\omega^i)$).
\end{itemize}

We can now prove by induction on the structure of $\psi$ that (a) for
all formulas $\psi$, $i = 1, \ldots, n$, and 
states $\omega' \in \Omega^i$, we have that 
$(M^i,\omega') \sat \psi$ iff $(M^*,\omega') \sat \psi$; 
(b)  for all formulas $\psi \in \L^4_{i+}$, $1 \le i, j \le n$,
$(M^i,\omega^i) \sat \psi$ iff $(M^*,\omega^j) \sat \psi$ (again,
here it is important that $\PR_i^*(\omega^j) = 0$ for $j = 1,\ldots,
n$).  Note that part (b) implies that the states $\omega^1, \ldots,
\omega^n$ satisfy the same formulas in $M^*$.  It easily follows that 
$(M^*,\omega^i) \sat \phi_1 \land \ldots \land \phi_n$ for $i =
1,\ldots, n$. \eprf

We can now prove the theorem.  
Again, let $X^k_j$ be the strategies for player $j$ that survive $k$
rounds of iterated deletion of weakly dominated strategies.
To see that (a) implies (b), suppose that 
$\sigma_i \in X^{k+1}_i$.  By Proposition~\ref{pro:Pearce}, there exists
a distribution $\mu_i$ whose support is $X^k_{-i}$ such that $\sigma_i$
is a best response to $\mu_i$.  By the induction hypothesis, for each
strategy profile $\tau_{-i} \in X^k_{-i}$, and all $j \ne i$, the
formula $\play_{j}(\tau_{j}) \land F^0_{j}$ is satisfied in a measurable
structure.  By
Lemma~\ref{lem:paste1}, $\play_{-j}(\tau_{-j}) \land (\land_{j
\ne i} F^k_{j})$ is satisfied in a measurable structure.  Taking
$\phi$ to be  $\land_{j \ne i} F^k_{j}$, by Lemma~\ref{lem:paste},
there exists a measurable structure $M$ and state $\omega$ in $M$ such
that the marginal  of
$\PR_i(\omega)$ on $X^k_{-i}$ is $\mu_i$, $\strat_i(\omega)$ is
$\sigma_i$, and $(M,\omega) \sat B_i (\land_{j \ne i}F^k_{j}) \land
(\land_{\psi \in \L^3_{(-j)+}}  \Diamond(\psi \land (\land_{j \ne i} 
F^k_{j})) \rimp \dbj \psi)$.  It follows that $(M,\omega)
\sat \RAT_i$, 
and hence that $(M,\omega) \sat F^{k+1}_i$, as desired.

It is immediate that (b) implies (c). \eprf
}

\cor\label{cor:wd1}  The following are equivalent:
\begin{itemize}
\item[(a)] the strategy $\sigma$ for player $i$ survives iterated 
deletion of weakly dominated strategies;
\item[(b)] there is a measurable structure  $M$
that is appropriate for $\Gamma$ and a state $\omega$ such that
$\strat_i(\omega) = \sigma$ and $(M,\omega) \sat \dbi F^k_i$ for all $k
\ge 0$;
\item[(c)] there is a structure  $M$
that is appropriate for $\Gamma$ and a state $\omega$ such that
$\strat_i(\omega) = \sigma$ and $(M,\omega) \sat \dbi F^k_i$ for all $k
\ge 0$.  
\end{itemize}
\ecor

\fullv{
\prf The proof is essentially identical to that of
Corollary~\ref{cor:wd}, so is omitted here.
\eprf
}

\section{Complete and Canonical Structures}\label{sec:complete}

\subsection{Canonical Structures}
Intuitively, to check whether a formula is strongly admissible, 
and, more generally, to check if all agent $i$ knows is $\phi$, 
we want to start with a very rich structure $M$ that contains all possible
consistent sets of formulas, so that if  $\phi \land \psi$ is satisfied
at all, it is satisfied in that structure.   Motivated
by this intuition, Halpern and Lakemeyer \citeyear{HalLak94} worked in
the \emph{canonical} structure for their language, which contains a
state corresponding to every consistet set of formulas.    We do the
same thing here.

Define the canonical structure $M^c = (\Omega^c, \strat^c, \F^c, \PR_1^c,
\ldots, \PR^c_n)$ for $\L^4$ as follows:
\begin{itemize}
\item $\Omega^c = \{\omega_\Phi: \Phi$ is a realizable subset of
$\L^4(\Gamma\}$; 
\item $\strat^c(\omega_\Phi) = \vec{\sigma}$ iff $\play(\sigma) \in
\Phi$;
\item $\F^c = \{F_\phi: \phi \in \L^4\}$, where $F_\phi =
\{\omega_\Phi: \phi \in \Phi\}$;
\item $\Pr_i^c(\omega_\Phi)(F_\phi) = \inf\{\alpha: \pr_i(\phi) > \alpha
\in \Phi\}$.  
\end{itemize}

\lem $M^c$ is an appropriate measurable structure for $\Gamma$. \elem

\fullv{
\prf 
It is easy to see that $\F^c$ is a $\sigma$-algebra, since the complement
of $F_\phi$ is $F_{\neg \phi}$ and $\inter_{m=1}^\infty F_{\phi_i} =
F_{\land_{m=1}^\infty \phi_m}$.  
Given a strategy $\sigma$ for player $i$, $\intensionc{\sigma} = 
F_{\play_i(\sigma}) \in \F$.
Moreover, each realizable set $\Phi$ that includes $\play_i(\sigma)$
must also include $\pr_i(\play_i(\sigma)) = 1$, so that 
$\PR_i(\omega_\Phi)(\strat_i(\omega_\Phi)) = 
\PR_i(\omega_\Phi)(F_{\play_i(\strat_i(\omega_\Phi)}) = 1$.
Similarly, suppose that $\PR_i(\omega_\Phi) = \pi$.  Then 
$\{\omega \in \Omega^c: \PR_i(\omega) = \pi\} = \inter_{\phi \in \L^3} \inter_{\{
\alpha \in Q \inter [0,1]: \pi(\intensionc{\phi}) \ge \alpha\}} F_{\phi
\ge \alpha} \in \F^c$.  Moreover, if $\alpha \in Q \inter [0,1]$, then
$\pi(\intensionc{\phi}) \ge \alpha$ iff $\pr_i(\phi) \ge \alpha \in
\Phi$.  But if $\pr_i(\phi) \ge \alpha \in \Phi$, then
$\pr_i(\pr_i(\phi) \ge \alpha) = 1 \in \Phi$.  It easily follows that  
$\PR_i(\omega_\Phi)(\{\omega: \PR_i(\omega) = \pi\}) = 1$. 
Finally, the definition
of $\F^c$ guarantees that every set $\intensionc{\phi}$ is measurable
and that $\PR_i(\omega_\Phi)$ is indeed a probability distribution on
$(\Omega^c,\F^c)$.   
\eprf
}

The following result is the analogue of the standard ``truth lemma'' in
completeness proofs in modal logic.

\pro\label{pro:truth} For $\psi \in \L^4$, 
$(M^c,\omega_\Phi) \sat \psi$ iff $\psi \in \Phi$. \epro

\fullv{
\prf A straightforward induction on the structure of $\psi$.
\eprf
}

We have constructed a canonical structure for $\L^4$.  It follows easily
from Lemma~\ref{lem:realizable} that the canonical structure for $\L^3$
(where the states are realizable $\L^3$ sets) is isomorphic to $M^c$.
(In this case, the set $\F^c$ of measurable sets would be
the smallest $\sigma$-algebra containing $\intension{\phi}$ for $\phi
\in \L^3$.)  
Thus, the choice of $\L^3$ vs.~$\L^4$ does not play an important role
when constructing a canonical structure.

A strategy $\sigma_i$ for player $i$ survives iterated
deletion of weakly dominated strategies iff the $\L^4$ formula 
$\undominated(\sigma_i) = \play_i(\sigma_i) \land (\land_{k=1}^\infty \dbi
F^k_i)$ is satisfied at 
some state in the canonical structure.  But there are other structures
in which $\undominated(\sigma_i)$ is satisfied.   One way to get such a
struture is by essentially ``duplicating'' states in the canonical
structure.  The canonical structure can be
\emph{embedded} in a structure $M$ if, for all $\L^3$-realizable sets
$\Phi$, there is a state $\omega_\Phi$ in $M$ such that
$(M,\omega_\Phi)\sat \phi$ iff $\phi \in \Phi$.  Clearly
$\undominated(\sigma_i)$ is satisfied in any structure in which the
canonical structure can be embedded.

A structure in which the canonical structure can be embedded is in a
sense larger than the canonical structure.  But $\undominated(\sigma_i)$
can be satisfied in structures smaller than the canonical structure.
(Indeed, with some effort, we can show that it is satisfiable in a
structure with countably many states.)  There are two reasons for this.
The first is that to satisfy $\undominated(\sigma_i)$, there is no need
to consider a structure with states where all the players are
irrational.  It suffices to restrict to states where at least one player
is using a strategy that survives at least one round of iterated
deletion.  This is because players know their strategy; thus, in a state
where a strategy $\sigma_j$ for player $j$ is admissible, player $j$
must ascribe positive probability to all other strategies; however, in
those states, player $j$ still plays $\sigma_j$.  

A perhaps more interesting reason that we do not need the canonical
structure is our use of the language $\L_3$.   Strong admissibility
guarantees that player $j$ will ascribe positive probability to all
formulas $\phi$ consistent with rationality.  Since a finite conjunction
of formulas in $\L^3$ is also a formula in $\L^3$, player $j$ will
ascribe positive probability to all finite conjunctions of formulas
consistent with rationality.  But a 
state is characterized by a \emph{countable} conjunction of formulas.
Since $\L^3$ is not closed under countable conjunctions, a structure
that satisfies $\undominated(\sigma_i)$ may not have states
corresponding to all $L^3$-realizable sets of formulas.  If we had used
$\L^4$ instead of $\L^3$ in the definition of strong admissibility
(ignoring the issues raised earlier with using $\L^4$), then there would
be a state corresponding to every $\L^4$-realizable (equivalently,
$\L^3$-realizable) set of formulas.  Alternatively, if we consider
appropriate structures that are compact in a topology where all sets
definable by formulas (i.e., sets of the form $\intension{\phi}$, for
$\phi \in \L^3$) are closed (in which case they are also open, since
$\intension{\neg \phi}$ is the complement of $\intension{\phi}$), then
all states where at least one player is using a strategy that survives at
least one round of iterated deletion will be in the structure.

Although, as this discussion makes clear, the formula that characterizes
strong admissibility can be satisfied in structures quite different from
the canonical structure, the canonical structure does seem to be the
most appropriate setting for reasoning about statements involving ``all
agent $i$ knows'', which is at the heart of strong admissibility.  
Moreover, as we now show, canonical structures allow us to relate our
approach to that of BFK.

\subsection{Complete Structures}
BFK worked with complete structures.   We now want to show that $M^c$ is
complete, in the sense of BFK.  To make this precise, we need to recall
some notions from BFK (with some minor changes to be more consistent
with our notation).   

BFK considered what they called \emph{interactive probability
structures}. These can be viewed as a special case of probability
structures.  A \emph{BFK-like structure} (for a game $\Gamma$) is 
a probability structure $M = (\Omega, \strat, \F, \PR_1,\ldots, \PR_n)$
such that there exist spaces $T_1, \ldots, T_n$ (where $T_i$ can be
thought of as the \emph{type space} for player $i$) such that
\begin{itemize}
\item $\Omega$ is isomorphic to $\vec{\Sigma} \times \vec{T}$, via some
isomorphism $h$;
\item if $h(\omega) = \vec{\sigma} \times \vec{t}$, then 
\begin{itemize}
\item $\strat(\omega) = \vec{\sigma}$, 
\item taking $T_i(\omega) = t_i$ (i.e., the type of player $i$ in
$h(\omega)$ is $t_i$); the support of $\PR_i(\omega)$ is
contained in $\{\omega': \strat_i(\omega') = \sigma', T_i(\omega') =
t_i\}$, so that $\PR_i(\omega)$ induces a probability on
$\Sigma_{-i} \times T_{-i}$;
\item $\PR_i(\omega)$ depends only on $T_i(\omega)$, in the sense that
if $T_i(\omega) = T_i(\omega')$, then $\PR_i(\omega)$ and
$\PR_i(\omega')$ induce the same probability distribution on
$\Sigma_{-i} \times T_{-i}$.
\end{itemize}
\end{itemize}

A BFK-like structure $M$ whose state space is isomorphic to $\vec{\Sigma}
\times \vec{T}$ is \emph{complete} if, for every  for each distribution
$\mu_i$ over $\Sigma_{-i} 
\times T_{-i}$, there is a state $\omega$ in $M$ such that the
probability distribution on $\Sigma_{-i} \times T_{-i}$
induced by $\PR_i(\omega)$ is $\mu_i$.  

\pro $M^c$ is complete BFK-like structure. \epro

\fullv{
\prf A set $\Phi \subseteq
\L^3_i$ is \emph{$\L^3_i$-realizable} if there 
exists an appropriate structure $M$ for $\Gamma$ and state $\omega$ in
$M$ such that, for all formulas $\phi \in \L^3$,  $(M,\omega) \sat 
\phi$ iff $\phi \in \Phi$.  Take the type space $T_i$ to consist of all
$\L^3_i$-realizable sets of formulas.  There is an isomorphism  $h$
between $\Omega^c$ and $\vec{\Sigma} \times \vec{T}$, where $T_i(\omega)$
is the $i$-realizable type of
formulas of the form $\pr_i(\phi) \ge \alpha$ that are true at
$\omega$; that is, $h(\omega) = \strat(\omega) \times T_1(\omega) \times
\cdots \times T_n(\omega)$.  It follows easily from
Lemma~\ref{lem:paste1} that $h$ is a surjection. 
we can identify $\Omega^c$, the state space in the canonical structure,
with $\vec{\S} \times \vec{T}$.  

To prove that $M^c$ is complete, given a probability $\mu$ on
$\Sigma_{-i} \times  T_{-i}$, we must show that there is some state
$\omega$ in $M^c$ such that the probability induced by $\PR_i(\omega)$ on
$\Sigma_{-i} \times  T_{-i}$ is $\mu$.  
Let $M^{\mu} = (\Omega^{sigma,\mu}, \F^{\mu},
\strat^{\mu}, \PR_1^{\mu}, 
\ldots, \PR_n^{\mu})$, where 
$M^{\mu}$ are defined as follows:
\begin{itemize}
\item $\Omega^{\mu} = \Omega^c
\union \Sigma \times \{\mu\} \times
T_{-i}$; 
\item $\F^{\mu}$ is the smallest $\sigma$-algebra that contains
$\F^c$ and all sets of the form $\vec{\sigma} \times \{\mu\} \times
\intensioncp{\phi}$, and  $\intensioncp{\phi}$ consists of the all type
profiles $t_{-i}$ such that, for some state $\omega$ in $M^c$,
$(M^c,\phi) \sat \phi$ and $T_{-i}(\phi) = t_{-i}$;
\item $\strat^\mu(\omega) = \strat^c(\omega)$ for $\omega \in \Omega^c$,
and $\strat^\mu(\vec{\sigma} \times \{\mu\} \times \vec{t}) = \vec{\sigma}$;
\item $\PR_j^\mu(\omega) = \PR_j^c(\omega)$ for $\omega \in \Omega^c$,
$j = 1, \ldots, n$; for $j \ne i$, $\PR_j^\mu(\vec{\sigma} \times
\mu\times t_{-i}) =  \PR_j(\omega)$, where $\strat_j(\omega) = \sigma_j$
and $T_j(\omega) = t_j$ (this is well defined, since if
$\strat_j(\omega') = \sigma_j$ and $T_j(\omega') = t_j$, then
$\PR_j(\omega) = \PR_j(\omega')$; finally, 
$\PR_i^\mu(\vec{\sigma} \times \mu\times t_{-i})$ is a
distribution whose support is contained in $\{\sigma_i\} \times \Sigma_{-i}
\times \{\mu\} \times T_{-i}$, and $\PR_i^\mu(\vec{\sigma} \times
\mu\times t_{-i})(\vec{\sigma} \times \mu\times \intensioncp{\phi}) =
\mu(\intensioncp{\phi})$.  
\end{itemize}
Choose an arbitrary state $\omega \in \vec{\Sigma} \times \{\mu\} \times
T_{-i}$.  The construction of $M^\mu$ guarantees that for $\phi \in
\L^4_{(-i)+}$, $(M^\mu, \omega) \sat \pr_i(\phi) > \alpha$ iff
$\mu(\intensioncp{\phi}) > \alpha$.  
By the construction of $M^c$, there exists a state $\omega' \in 
\Omega^c$ such that $(M^c,\omega') \sat \psi$ iff $(M^\mu,\omega) \sat
\psi$.  Thus, the distribution on $\Sigma_{-i} \times T_{-i}$ induced by
$\PR_i(\omega)$ is $\mu$, as desired.  This shows that $M^c$ is
complete. 
\eprf
}

We now would like to show that every measurable complete BFK-like
structure is the canonical model.  This is not quite true because states
can be duplicated in an interactive structure.  This suggests that we
should try to show that the canonical structure can be embedded in every
measurable complete structure.  We 
can essentially show this, except that we need to restrict to
\emph{strongly measurable} complete structures, where a structure is
strongly measurable if it is measurable and the only measurable sets are
those defined by $\L_4$ formulas (or, equivalently, the set of
measurable sets is the smallest set that contains the sets defined by
$\L_3$ formulas).  
\fullv{
We explain where strong measurability is needed at
the end of the proof of the following theorem.}
\shortv{(We explain the need for strong measurability in the full paper.)}

\thm\label{thm:strong} If $M$ is a strongly measurable complete
BFK-like structure,  
then the canonical structure can be embedded in $M$.    
\ethm

\fullv{
\prf Suppose that $M$ is a strongly measurable complete BFK-like structure.
We can assume without loss of generality that the state space of $M$ has
the form $\vec{\Sigma} \times \vec{T}$.  To prove the result, we need
the following lemmas.

\lem\label{lem:depends} If $M$ is BFK-like, the truth of a formula $\phi
\in \L^4_i$ at a state $\omega$ in $M$ depends only on $i$'s type; 
That is,  if $T_i(\omega) = T_i(\omega')$, then $(M,\omega) \sat \phi$
iff $(M,\omega') \sat \phi$.  Similarly, the truth of a formula in
$\L_{i+}$ in $\omega$ depends only on $\strat_i(\omega)$ and
$T_i(\omega)$, and the truth of a formula in
$\L^4_{i+}$  in $\omega$  depends only on $T_{-i}(\omega)$.
\elem

\prf  A straightforward induction on structure. \eprf

Define a \emph{basic formula} to be one of the form $\psi_1\land \ldots
\land \psi_n$, where $\psi_i \in \L^3_{i+}$ for $i = 1,\ldots,n$.

\lem\label{lem:basic} Every formula in $\L^3$ is equivalent to a finite
disjunction of basic formulas.  
\elem

\prf A straightforward induction on structure. \eprf

\lem\label{lem:basic1} Every formula in $\L^3_{i+}$ is equivalent to a 
disjunction of formulas of the form 
\begin{equation}\label{eq1}
\play_i(\sigma) \land (\neg) RAT_i
\land (\neg) \pr_i(\phi_1) > \alpha_1 \land \ldots  \land
(\neg)\pr_i(\phi_m) > \alpha_m \land (\neg) \pr_i(\psi_1) \ge \beta_1
\land \ldots  \land (\neg)\pr_i(\psi_{m'}) \ge \beta_{m'},
\end{equation}
where $\phi_1, \ldots, \phi_m, \psi_1, \ldots, \psi_{m'} \in
 \L^3_{(-i)+}$ and the
``($\neg$)'' indicates that the presence of negation is optional.
\elem

\prf A straightforward induction on the structure of formulas, using the
observation that $\neg\play_i(\sigma)$ is equivalent to $\lor_{\{sigma'
\in \Sigma_i: \sigma' \ne \sigma\}} \play_i(\sigma')$. \eprf

\lem\label{lem:satisfiable} If $\phi \in \L^3$ is satisfiable, then
$\intension{\phi} \ne \emptyset$.  \elem

\prf By Lemma~\ref{lem:basic}, it suffices to prove the result for the
case that $\phi$ is a basic formula.  By Lemma~\ref{lem:basic1}, it
suffices to assume that the the ``$i$-component'' of the basic formula
is a conjunction.  We now prove the result by induction on the depth 
of nesting of the modal operator $\pr_i$ in $\phi$.  (Formally, define
$D(\psi)$, the depth of nesting of $\pr_i$'s in $\psi$, by induction on
the structure of $\psi$.  if $\psi$ has the form $\play_j(\sigma)$,
$\RAT_j$, or $\true$, then $D(\psi) = 0$; $D(\neg \psi) = D(\psi)$; 
$D(\psi_1 \land \psi_2) = \max(D(\psi_1),D(\psi_2))$; and $D(\pr_i(\psi)
> \alpha) = D(\pr_i(\psi) \ge \alpha) = 1 + D(\psi)$.)  
Because the state space $\Omega$ of $M$ is essentially a product space,
by Lemma~\ref{lem:depends}, it suffices to prove the result for formulas
in $\L^3_{(i)+}$.   It is clear that $\phi$ possibly puts constraints on
what strategy $i$ is using, the probability of strategy profiles in
$\Sigma_{-i}$, and the probability of formulas that appear in the scope
of $\pr_i$ in $\phi$.  If $M' = (\Omega',\strat',\F',\PR_1',\ldots,
\PR_n')$ is a structure and $\omega' \in \Omega'$, 
then $(M',\omega') \sat \phi$ iff $\strat'_i(\omega')$ and
$\PR'_i(\omega')$ satisfies these constraints.  (We leave it to the reader
to formalize this somewhat informal claim.)  
By the induction hypothesis, each formula in the scope
of $\pr_i$ in $\phi$ that is assigned positive probability by
$\PR_i(\omega')$ is satisfied in $M$.  Since $M$ is complete and
measurable, there is a
state $\omega$ in $M$ such that $\strat_i(\omega) = \strat'_i(\omega')$
and $\PR_i(\omega)$ places the same constraints on formulas that appear
in $\phi$ as $\PR_i$.  We must have $(M,\omega) \sat \phi$.   
\eprf  

Returning to the proof of the theorem, suppose that $M =
(\Omega,\strat,\F, \PR_1, \ldots, \PR_n)$.  
Given a state $\omega \in \Omega^c$, we claim that there must be a state
$\omega'$ in $M$ such that $\strat(\omega') = \strat^c(\omega)$ and, for
all $i = 1, \ldots, n$, $\PR_i^c(\omega)(\intensionc{\psi}) =
\PR_i(\omega')(\intension{\psi})$.  to show this, because of $\Omega$ is a
product space, and $\PR_i(\omega')$ depends only on $T_i(\omega')$, it
suffices to show that, for each $i$, there exists a state $\omega_i$ in
$M$ such that, for each $i$, $\PR_i^c(\omega)(\intensionc{\psi}) =
\PR_i(\omega_i)(\intension{\psi})$.  By Lemma~\ref{lem:satisfiable},
if $\intensionc{\psi} \ne \emptyset$, then $\intension{\psi} \ne
\emptyset$.  Thus, the existence of $\omega_i$ follows from
the assumption that $M$ is complete and strongly measurable.

Roughly speaking, 
To understand the need for strong measurability here, note that even
without strong measurability, the argument above tells us that 
there exists an appropriate measure defined on 
sets of the form $\intension{\phi}$ for $\phi$ in $\L^3_{(-i)+}$.  We can
easily extend $\mu$ to a measure $\mu'$ on sets of the form
$\intension{\phi}$ for $\phi$ in $\L^4_{(-i)+}$.  However, if the set
$\F$ of measurable sets in $M$ is much richer than the sets definable by
$\L^4$ formulas, it is not 
clear that we can extend $\mu'$ to a measure on all of $\F$.  In
general, a countably additive measure defined on a subalgebra of  a set
$\F$ of measurable sets cannot be extended to $\F$.  For example, it is
known that, under the continuum hypothesis, 
Lebesgue measure defined on the Borel sets cannot be extended to all
subsets of $[0,1]$ \cite{Ulam30}; see \cite{KT64} for further
discussion).  Strong measurability allows us to avoid this problem.
\eprf
}

\bibliographystyle{chicagor}
\bibliography{z,joe}
\end{document}

%% file: defn.tex
\newtheorem{THEOREM}{Theorem}[section]
\newenvironment{theorem}{\begin{THEOREM} \hspace{-.85em} {\bf :} }%
                        {\end{THEOREM}}
\newtheorem{LEMMA}[THEOREM]{Lemma}
\newenvironment{lemma}{\begin{LEMMA} \hspace{-.85em} {\bf :} }%
                      {\end{LEMMA}}
\newtheorem{COROLLARY}[THEOREM]{Corollary}
\newenvironment{corollary}{\begin{COROLLARY} \hspace{-.85em} {\bf :} }%
                          {\end{COROLLARY}}
\newtheorem{PROPOSITION}[THEOREM]{Proposition}
\newenvironment{proposition}{\begin{PROPOSITION} \hspace{-.85em} {\bf :} }%
                            {\end{PROPOSITION}}
\newtheorem{DEFINITION}[THEOREM]{Definition}
\newenvironment{definition}{\begin{DEFINITION} \hspace{-.85em} {\bf :} \rm}%
                            {\end{DEFINITION}}
\newtheorem{CLAIM}[THEOREM]{Claim}
\newenvironment{claim}{\begin{CLAIM} \hspace{-.85em} {\bf :} \rm}%
                            {\end{CLAIM}}
\newtheorem{EXAMPLE}[THEOREM]{Example}
\newenvironment{example}{\begin{EXAMPLE} \hspace{-.85em} {\bf :} \rm}%
                            {\end{EXAMPLE}}
\newtheorem{REMARK}[THEOREM]{Remark}
\newenvironment{remark}{\begin{REMARK} \hspace{-.85em} {\bf :} \rm}%
                            {\end{REMARK}}
\newcommand{\thm}{\begin{theorem}}
\newcommand{\lem}{\begin{lemma}}
\newcommand{\pro}{\begin{proposition}}
\newcommand{\dfn}{\begin{definition}}
\newcommand{\rem}{\begin{remark}}
\newcommand{\xam}{\begin{example}}
\newcommand{\cor}{\begin{corollary}}
\newcommand{\prf}{\noindent{\bf Proof:} }
\newcommand{\ethm}{\end{theorem}}
\newcommand{\elem}{\end{lemma}}
\newcommand{\epro}{\end{proposition}}
\newcommand{\edfn}{\bbox\end{definition}}
\newcommand{\erem}{\bbox\end{remark}}
\newcommand{\exam}{\bbox\end{example}}
\newcommand{\ecor}{\end{corollary}}
\newcommand{\eprf}{\bbox\vspace{0.1in}}
\newcommand{\beqn}{\begin{equation}}
\newcommand{\eeqn}{\end{equation}}
\newcommand{\bbox}{\vrule height7pt width4pt depth1pt}

\newcommand{\clm}{\begin{claim}}
\newcommand{\eclm}{\end{claim}}
\newcommand{\sat}{\models}

\newcommand{\rimp}{\Rightarrow}
\newcommand{\union}{\cup}
\newcommand{\inter}{\cap}
\renewcommand{\phi}{\varphi}
\newcommand{\F}{{\cal F}}

\newcommand{\Z}{{\cal Z}}

\newcommand{\ol}{\setlength{\itemsep}{0pt}\begin{enumerate}}
\newcommand{\eol}{\end{enumerate}\setlength{\itemsep}{-\parsep}}
\newcommand{\ul}{\setlength{\itemsep}{0pt}\begin{itemize}}
\newcommand{\dl}{\setlength{\itemsep}{0pt}\begin{description}}
\newcommand{\edl}{\end{description}\setlength{\itemsep}{-\parsep}}
\newcommand{\eul}{\end{itemize}\setlength{\itemsep}{-\parsep}}

\newcommand{\true}{\mbox{{\it true}}}

\newcommand{\commentout}[1]{}

\newcommand{\bi}{\begin{itemize}}
\newcommand{\ei}{\end{itemize}}
\newcommand{\be}{\begin{enumerate}}
\newcommand{\ee}{\end{enumerate}}